%% file: arxiv.tex
\documentclass[keeplastbox, letterpaper, 10 pt, conference]{ieeeconf}  % Comment this line out if you need a4paper

\IEEEoverridecommandlockouts                              % This command is only needed if 
\makeatletter
\let\NAT@parse\undefined
\makeatother

\usepackage{graphics}  % for pdf, bitmapped graphics files
\usepackage{graphicx}

\usepackage{bm}        % bold math (greek lowercase)
\usepackage{amsmath}   
\usepackage{amssymb}   

\usepackage{balance}

\usepackage{float}
\usepackage{makecell}
\usepackage[normalem]{ulem}
\usepackage{hyperref}
\usepackage{color}

\usepackage{amsfonts}
\usepackage{algorithm,algpseudocode}

\usepackage{gensymb}
\usepackage[tbtags]{mathtools}

\usepackage[pdftex,dvipsnames]{xcolor}

\usepackage{todonotes}
\usepackage{units}

\usepackage{tikz}
\usepackage{textcomp}

\newcommand{\norm}[1]{\left\lVert#1\right\rVert}

\usepackage[comma,numbers,compress]{natbib}

\usepackage[firstpage=true]{background}

\SetBgContents{\textbf{Accepted final version.} In \textit{2018 IEEE-RAS 18th International Conference on Humanoid Robots (Humanoids)}}
\SetBgScale{1}
\SetBgColor{black}
\SetBgAngle{0}
\SetBgOpacity{1}
\SetBgPosition{current page.south}
\SetBgVshift{1cm}

\title{\LARGE \bf
Learning Postural Synergies for Categorical Grasping through\\ Shape Space Registration}

\author{Diego Rodriguez$^{1}$, Antonio Di Guardo$^{2}$, Antonio Frisoli$^{2}$, and Sven Behnke$^{1}$% <-this % stops a space
\thanks{$^{1}$Diego Rodriguez and Sven Behnke are with the Autonomous Intelligent Systems (AIS) Group, Computer Science Institute VI, University of Bonn, Germany,
        {\tt\small \{rodriguez, behnke\}@ais.uni-bonn.de} }%
\thanks{$^{2}$Antonio Di Guardo and Antonio Frisoli are with the PERCRO Laboratory, TeCIP Institute, Scuola Superiore Sant'Anna, Pisa, Italy, {\ttfamily\small antonio.\{diguardo, frisoli\}@santannapisa.it } }
\thanks{This work was supported by the European Union's Horizon 2020 Programme under Grant Agreement 644839 (CENTAURO) and the German Research Foundation (DFG) under the grant BE 2556/12 ALROMA in priority programme SPP 1527 Autonomous Learning.}
}%

\begin{document}

\maketitle
\thispagestyle{empty}
\pagestyle{empty}

%%%%%%%%%%%%%%%%%%%%%%%%%%%%%%%%%%%%%%%%%%%%%%%%%%%%%%%%%%%%%%%%%%%%%%%%%%%%%%%%
\begin{abstract}
	Every time a person encounters an object with a given degree of familiarity, he/she immediately knows how to grasp it.
	Adaptation of the movement of the hand according to the object geometry happens effortlessly because of the accumulated knowledge of previous experiences grasping similar objects.
	In this paper, we present a novel method for inferring grasp configurations based on the object shape.
	Grasping knowledge is gathered in a synergy space of the robotic hand built by following a human grasping taxonomy.
	The synergy space is constructed through human demonstrations employing a exoskeleton that provides force feedback,
	which provides the advantage of evaluating the quality of the grasp.
	The shape descriptor is obtained by means of a categorical non-rigid registration that encodes typical intra-class variations.
	This approach is especially suitable for on-line scenarios where only a portion of the object's surface is observable.
	This method is demonstrated through simulation and real robot experiments by grasping objects never seen before by the robot.
\end{abstract}

\input{introduction}
\input{related_work}
\input{exo_skeleton}
\input{post_synergies}
\input{shape_space_reg}
\input{learn_synergies}
\input{experiments}
\input{conclusion}

\balance

\bibliographystyle{IEEEtranN}
\bibliography{humanoids_synergies}

\end{document}

%% file: introduction.tex
%%%%%%%%%%%%%%%%%%%%%%%%%%%%%%%%%%%%%%%%%%%%%%%%%%%%%%%%%%%%%%%%%%%%%%%%%%%%%%%%
\section{Introduction}
\label{sec:introduction}
The object geometry is a key element for a successful grasp.
Based on geometrical variations, humans are able to transfer previous knowledge of similar objects to new observed instances and to perform new grasps.
In this paper, we aim to provide this capability to robots, i.e., grasp adaptation according to the object shape.
We do this by inferring a postural synergy from a shape description of the object.
This shape description is obtained through a non-rigid category-based registration that captures geometrical object variations inside a category.
This descriptor resides in a low dimensional shape space of the category.

To describe the grasp configuration of the robotic hand, we use postural synergies because of its lower dimensionality compared to the number of Degrees of Freedom (DoFs) of the hand.
In this manner, we reduce the output dimensionality of the learning approach presented here.
The synergy space is constructed following a human grasping taxonomy \citep{feix2016}.
However, we do not rely on any visual sensory data or any human-to-robot mapping, 
but we directly acquire the joint space configuration of the robotic hand through a teleoperated exoskeleton.
Thus, we avoid errors coming from camera calibrations and mappings to the robotic kinematics.
In addition, the exoskeleton provides the user with force feedback which serves to assess the grasps qualitatively.
Moreover, the complexity and consequently the required time building the synergy space is considerably reduced.

The approach presented here is aimed for on-line grasping scenarios. 
Our shape space registration is able to infer the shape descriptor from a \emph{single view} of the object coming from RGBD sensors,
This is possible because our shape space registration is able to reconstruct to a certain extent partially occluded parts of the objects.

\begin{figure}[t]
	\centering
	\includegraphics[width=0.8\linewidth]{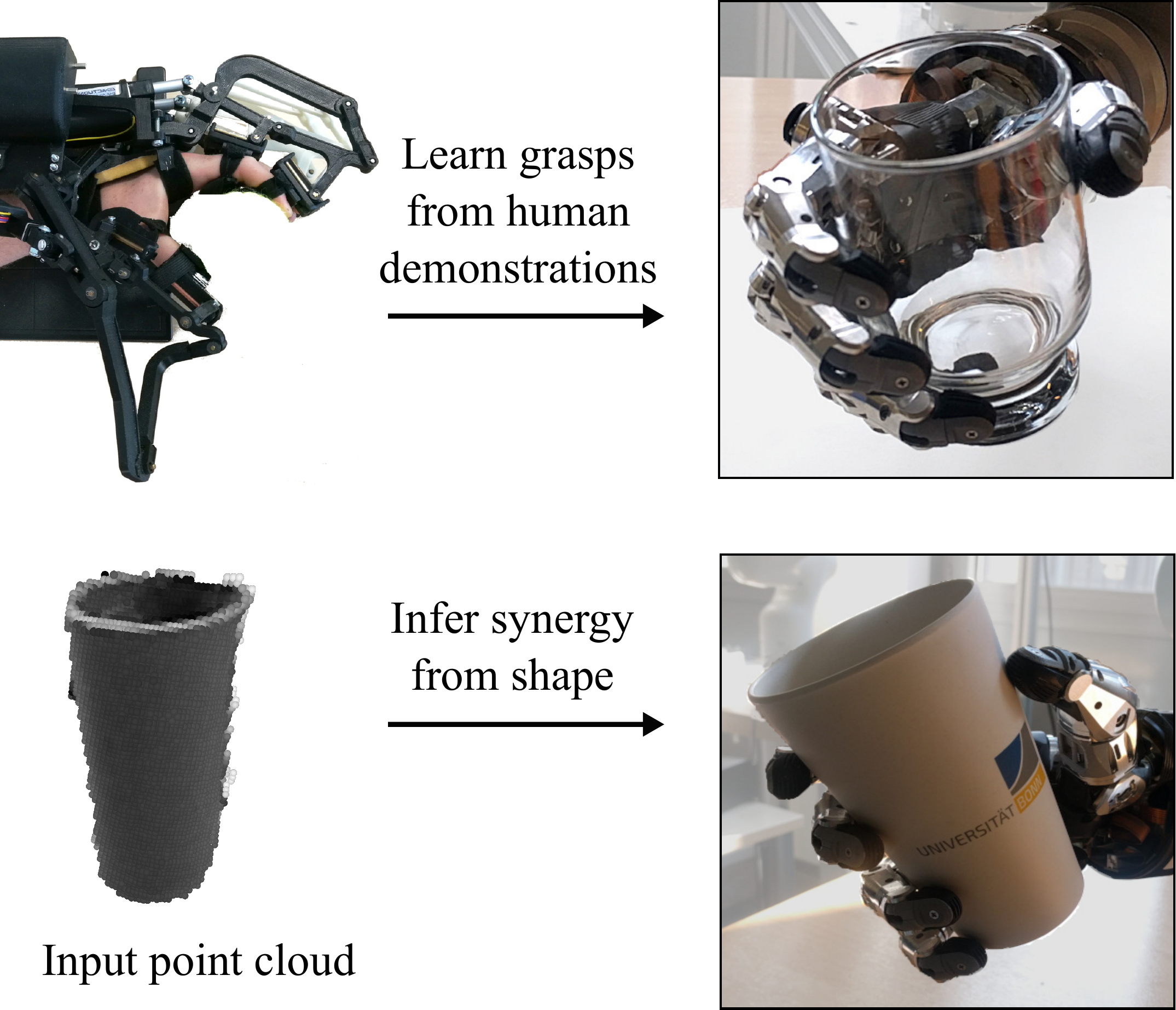} 
	\caption{Grasping configurations of objects in a category are inferred from a description of their shape.
		Grasping knowledge is encoded in a synergy space and a grasping learner through human demonstrations using an exoskeleton.}
	\label{fig:overview}
\end{figure}

The main contributions of this paper are: the inference of grasp configurations from the extrinsic object geometry based on a category-based shape space
and the generation for the first time, to the best of the author's knowledge, of a grasping postural synergy space using force feedback provided by a hand exoskeleton (Fig.\ref{fig:overview}).
A video illustrating our approach is available online \footnote{\url{www.ais.uni-bonn.de/videos/Humanoids_2018_Rodriguez}}.

%% file: related_work.tex
\section{Related Work}
\label{sec:related_work}
\subsection{Shape Space Registration}
Standard non-rigid registration methods such as conformal maps\citep{zeng2010dense}, 
thin-plate splines~\cite{brown2007global} and the coherent point drift~\citep{myronenko2010point} are able to quantify deformations between two objects, 
but they do not possess any notion of category-level features,
which can be exploited for reconstruction in on-line scenarios.
\citet{burghard2013compact} proposed a shape space of strongly varying geometry to establish dense correspondences.
This method, however, does not achieve good results with noisy data or partial views of the objects.
A shape manifold that models intra-category shape variances and is robust against noisy or occluded parts was presented by \citet{engelmann2016joint}. 
However, this method does not provide any deformation field or correspondences.

To solve these problems, we proposed in~\cite{Rodriguez2018a} a novel non-rigid registration method that incorporates category-level information and is able to register partially-occluded instances using a single capture of the object.
This approach combines the Coherent Point Drift (CPD) with subspaces methods to create a shape (latent) space that encodes typical geometrical variations inside a category.
This method has been applied to transfer control poses for approaching the objects to grasp~\cite{Klamt2018, Pavlichenko2018}, 
and to accumulate experiences on the motion for grasping different objects into a canonical model~\cite{Rodriguez2018b}.
In this paper, we mainly concentrate on the configuration of the hand,
i.e., on how the hand configuration changes according to the shape of the objects.
\subsection{Synergy-based Grasping}
Postural synergies have been widely accepted in the robotics community as a grasp representation for control and planning mainly because of their lower dimensionality compared with the number of DoFs of the robotic hands~\citep{ciocarlie, ficuciello2017, bernardino2013, ficuciello2016}.
In order to generate the synergy space, an anthropomorphic taxonomy is often followed~\cite{feix2016}.
Some approaches use visual sensory data to acquire human grasp poses to posteriorly map them to the kinematics of the robotic hand~\cite{ficuciello2017, amor2012generalization}.
However, errors coming from the visual system or from the human-to-robot kinematics mapping severely affect this kind of approaches.
\citet{bernardino2013} overcomes this limitation by acquiring directly the joint position of the robotic hand through a data glove.
Our approach enriches this data acquisition by employing teleoperation force feedback. 
In this scenario, the human user can reach stable grasping solutions by relying on both visual and force feedback.
\subsection{Learning Grasp Synthesis based on Object Shapes}
\citet{ekvall} infer approaching vectors based on shape primitives and human demonstrations obtained by data gloves.
\citet{ficuciello2016} are able to adapt postural synergies in a reinforcement learning manner based on a force-closure cost function.
Later, \citet{ficuciello2017} infers synergy values from a by-user-given basic description (diameter, length and height) of the objects using a neural network.
Our approach, on the other hand, infers a more complex shape description of the objects autonomously by making use of our shape space registration.
A similar approach as the one proposed in this paper is described by \citet{faria}. 
There, objects represented as point clouds are segmented into parts and represented as superquadrics parameters.
Based on these parameters synergies vectors are inferred in a Bayesian fashion.
However, unlike \cite{faria} we ensure that the objects are grasped in a functional way and the model is reconstructed due to the knowledge residing in the latent space.

%% file: exo_skeleton.tex
\section{Hand Teleoperation}
\label{sec:exo_skeleton}
\begin{figure}
	\centering
	\includegraphics[width=0.65\linewidth]{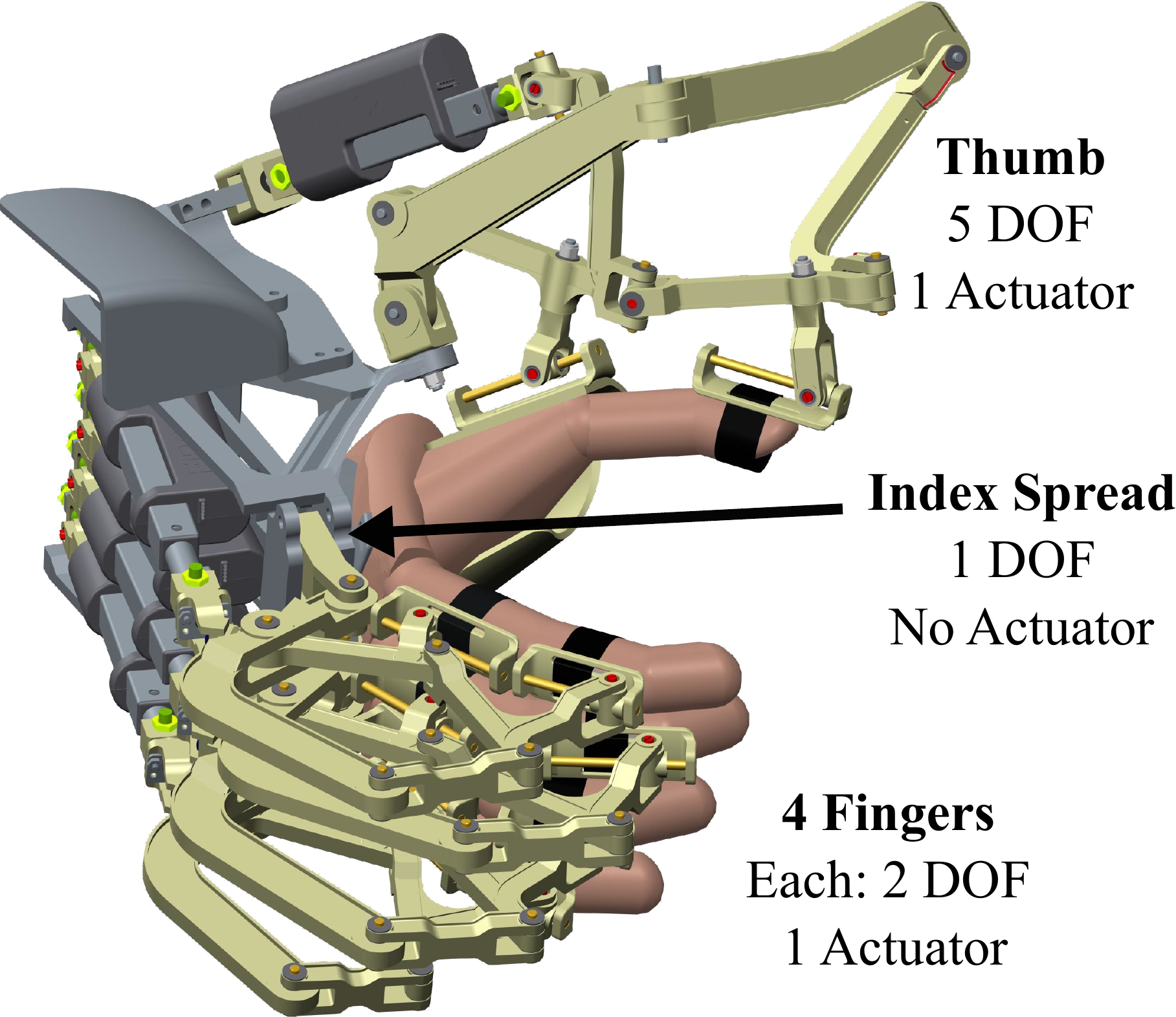}
	\caption{3D rendering of the \textit{Underactuated Hand Exoskeleton} U-HEx. 
		In total the exoskeleton has 14 DoFs.}
	\label{fig:U-HEx}
\end{figure}
The control of multi-fingered robotic hands is a complex problem because of its high number of independent variables. 
In this scenario, the design symmetry between human and robotic hand allows an operator to teleoperate the manipulator in a very natural way. 
In particular, bilateral telemanipulation requires the adoption of a haptic device such as the hand exoskeleton. 
\begin{figure}[b]
	\centering
	\includegraphics[width=\linewidth]{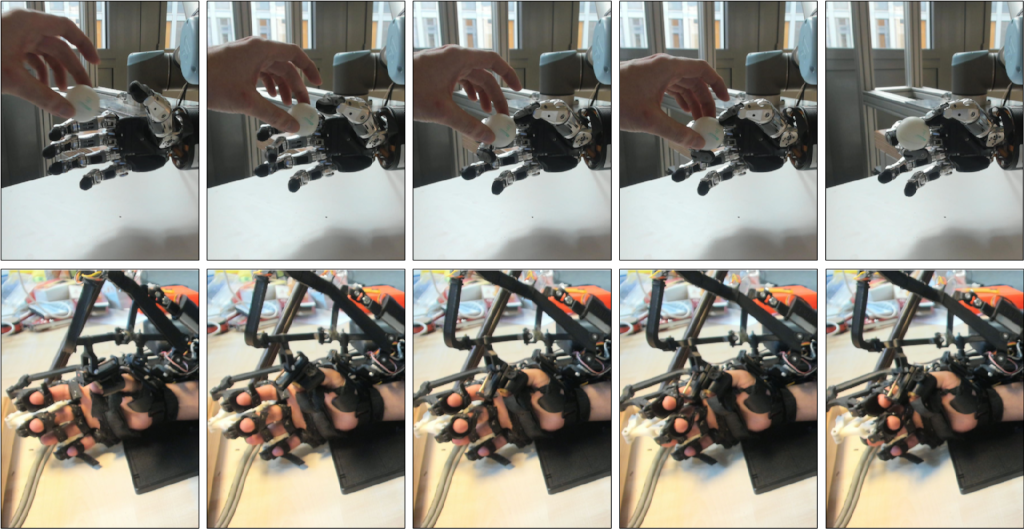}
	\caption{Snapshots of U-HEx and Schunk Hand during force feedback teleoperation.}
	\label{fig:teleoperation}
\end{figure}
\begin{figure*}
	\centering
	\includegraphics[width=\linewidth]{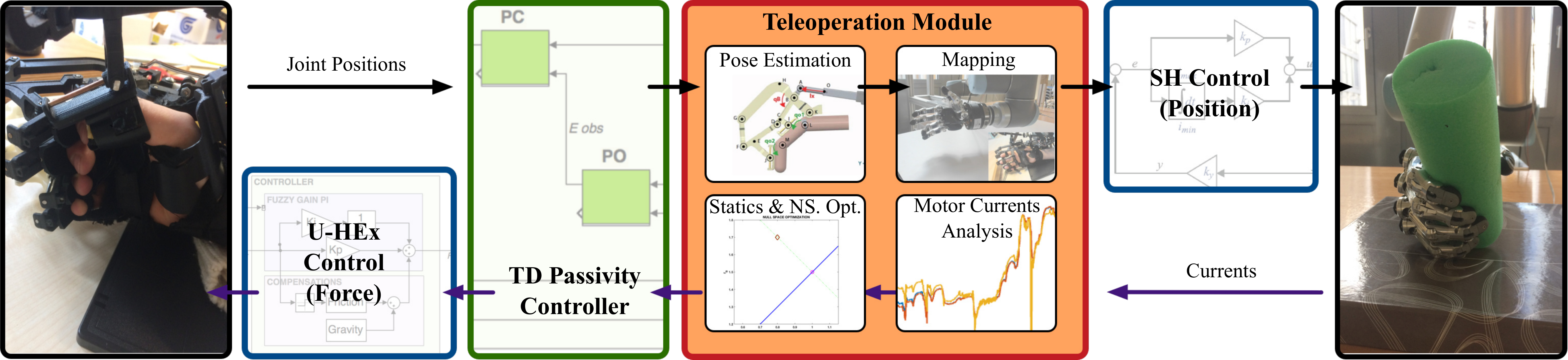}
	\caption{Overview of Schunk Hand (SH) position-force teleoperation architecture. The teleoperation module runs at 100 Hz, while low level controllers run at 1 KHz.}
	\label{fig:control_highLevel}
\end{figure*}
The data acquisition for generating the synergy space was carried out using the U-HEx, a novel underactuated hand exoskeleton developed by the PERCRO lab of Scuola Superiore Sant'Anna.
The device is composed of five independent parallel kinematics attached to a base fixable to the operator hand-back (Fig.~\ref{fig:U-HEx}). 
The five DoFs of the thumb exoskeleton allow identifying all the joint values associated with the human movements~\citep{gabardi2018design}.
The index finger possesses three DoFs, all the other exoskeleton's fingers have only two DoFs,
making observable only the \textit{metacarpal-proximal joint} (MCP) and the \textit{proximal-interphalanx joint} (PIP)~\citep{sarac2017design}. 
The exoskeleton has 14 DoFs but only five actuators, one actuator for each finger. 
Parallelism and underactuation ensure lightweight, minimal bulkiness and high adaptability to different hand sizes without mechanical adjustments. 
Kinesthetic forces are transmitted through two human-exoskeleton contact points for each finger.

The kinematic correspondence between the human and anthropomorphic hands is performed using a \textit{position-force} bilateral teleoperation scheme. 
An initial calibration phase is required as described in \cite{di2018sensitivity}.
In each execution cycle, the exoskeleton identifies operator finger joints angles that are directly mapped to the corresponding joint angles of the robotic hand exploiting the symmetry between the two kinematics. 
Whenever the symmetry is not present (i.e. in the case of the thumb), motor commands are defined as linear combinations between human joint (Fig. \ref{fig:control_highLevel}).
Linear combination coefficients are empirically set to match the master and slave workspaces. 

Force reflection channel is performed as follows. 
First, remote interaction forces are observed in terms of joint torques (linearly mapped from motor currents).
Next, each joint torque is scaled and applied to the corresponding operator hand joint.
In absence of communication delays, stability is empirically enhanced employing small values of force reflection gains and by means of the one port passivity at the exoskeleton motors level \cite{hannaford2002time}. 
The underactuation problem is solved through a null-space optimization method as described in \citep{sarac2018rendering}. 
The proposed teleoperation architecture not only allows the operator to grasp objects with very different sizes and shapes, but also to modulate interaction forces making possible the grasp of fragile and deformable objects (Fig. \ref{fig:teleoperation}).

%% file: post_synergies.tex
\section{Postural Synergies}
\label{sec:synergies}

\begin{figure*}
	\centering
	\footnotesize
	\includegraphics[width=0.7\linewidth]{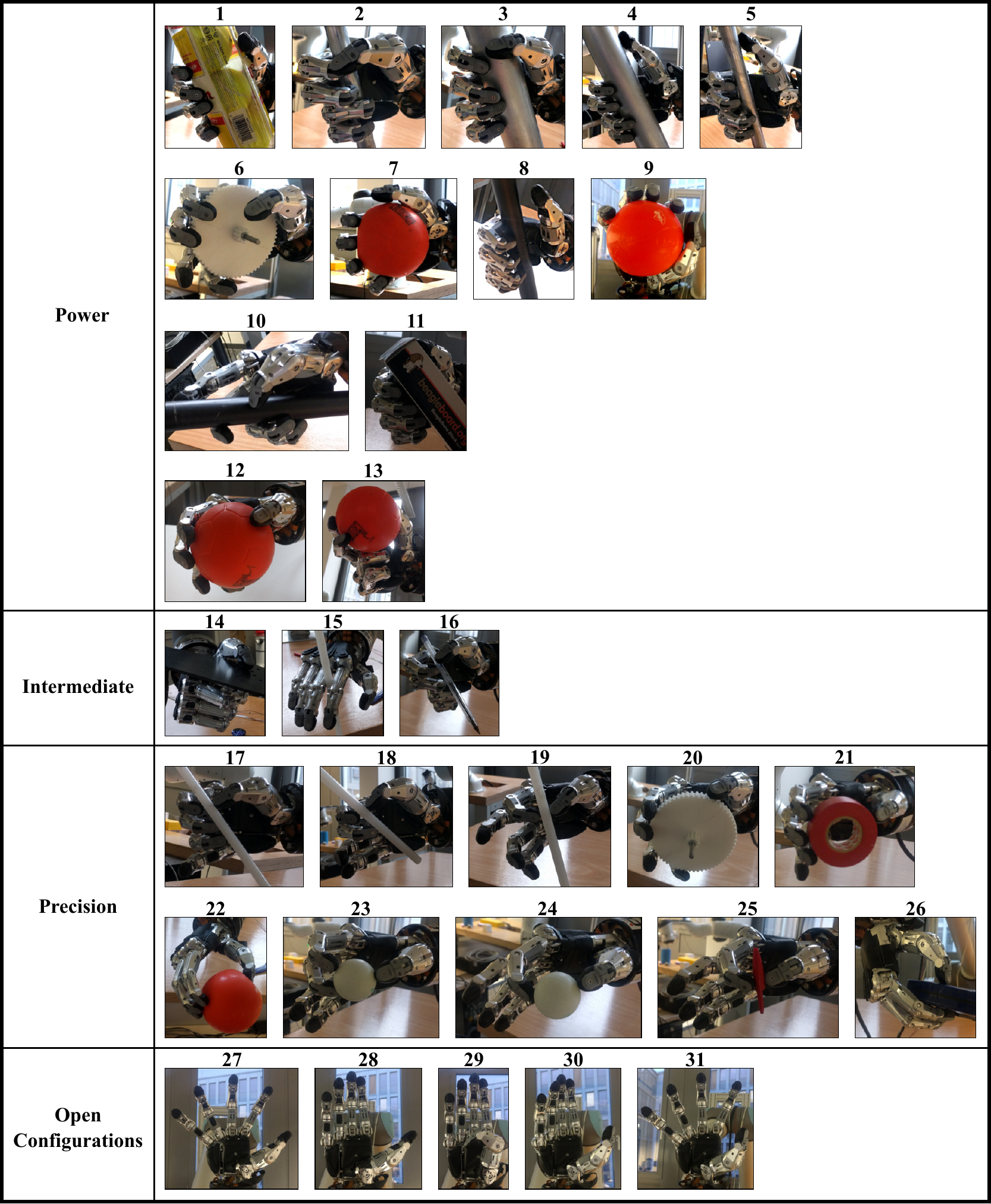} 
	\caption{Grasped objects used to calculate the synergy space. 
		All grasps are categorized in four groups: \textit{power}, \textit{intermediate}, \textit{precision} and \textit{open configurations}.}
	\label{fig:post_synergies}
\end{figure*}

We used the anthropomorphic multi-fingered Schunk hand as the robotic platform for the approaches presented here.
The hand has 20 joints but only 9 of them are fully actuated, the other 11 are coupled or mimic joints.
The DoFs are distributed as follows: thumb (2), index finger (2), middle finger (2), ring finger (1), pinky (1) and finger spread (1).
One limitation of the hand kinematics is the spread movements of the finger which is controlled only by one motor.
This limitation imposes a hard constraint on the Cartesian position of the fingers,
because for 4 different target poses (index, middle, ring and pinky fingers) only one can be guaranteed if it is contained in the workspace of the finger.
Because of this limitation on the hand kinematics and the fact that each finger has one or maximum two flexion DoFs,
a mapping coming from real human grasping joint angles to the joint space of the robotic hand will not exploit all the grasping capabilities of the robotic hand.
For this reason, we use the exoskeleton presented in Sec. \ref{sec:exo_skeleton} for controlling the hand to the desired grasping configurations.

We performed $n=$31 grasps following the human grasping taxonomy presented in \cite{feix2016},
in which 33 grasps have been grouped into \textit{power}, \textit{intermediate} and \textit{precision}.
With the Schunk Hand, we were able to reproduce 26 of them (Fig.~\ref{fig:post_synergies}).
The missing seven grasp configurations were not feasible due to kinematics limitations.
Moreover, five open configurations were included.

For each grasp, the join configuration $\mathbf{q}$ given by the robot hand was recorded and assembled into a matrix $\mathbf{A}\!=\!\{\mathbf{q}_1^T\!-\!\mathbf{\bar{q}}^T,\dots,\mathbf{q}_n^T\!-\!\mathbf{\bar{q}}^T\}\!\in\!\mathbb{R}^{n\times q}$, where $\mathbf{\bar{q}}$ represents the mean joint position.
By decomposing the symmetric positive matrix $\mathbf{A}^T\mathbf{A}\!=\!\mathbf{Q}\mathbf{\Lambda}\mathbf{Q}^T$ into its eigenvalues and eigenvectors
and taking the $l$ eigenvectors with the highest eigenvalues, 
the vector base $\mathbf{L}\!\in\!\mathbb{R}^{q\times l}$ of the $l$-dimensional synergy subspace is calculated.
Thus, the synergy $\mathbf{s}$ corresponding to $\mathbf{q}$ can be expressed as:
\begin{equation}
\label{eq:to_synergy}
\mathbf{s} = \mathbf{L}^T(\mathbf{q}-\mathbf{\bar{q}}),
\end{equation}
while the inverse transformation is described by:
\begin{equation}
\label{eq:to_q}
\mathbf{q} = \mathbf{\bar{q}} + \mathbf{L}\mathbf{s} .
\end{equation}

For the synergy space of the Schunk Hand, 
the explained variance of the two principal components equals 78\% while employing three components is 88\%.
These results are comparable with the total explained variance observed in humans: 84\% and 90\% for two and three components, respectively, where 15 joints were recorded \cite{Santello1998}.
These results are also comparable with the UB hand IV that possesses 15 DoFs; 
the explained variance equals 75\% and 90\%, for two and three components, respectively \cite{Ficuciello2014}.

\subsection{Inverse Kinematics in Synergy Space}
As occurred in joint space, in the synergy space, the configuration of the hand might result in a self-collision.
An inverse kinematics solver is then proposed to generate collision-free grasps.
The solver works directly in the synergy subspace avoiding to perform operations in higher dimensions, e.g., Jacobian computations in joint space.
The final synergy pose is computed iteratively, such that:
\begin{equation}
	\mathbf{s}_{i+1}=\mathbf{s}_i+\Delta \mathbf{s} .
\end{equation}

In each iteration, the following quadratic programming problem is solved:
\begin{equation}
	\begin{aligned}
	& \underset{\mathbf{\dot{s}}}{\text{minimize}}
	& & \frac{1}{2}\mathbf{\dot{s}}^T\mathbf{J}^T\mathbf{J}\mathbf{\dot{s}}+\mathbf{r}^T\mathbf{J}\mathbf{\dot{s}} \\
	& \text{subject to}
	& & \mathbf{G}_k\mathbf{\dot{s}} \leq \mathbf{h}_k
	\end{aligned}
\end{equation}
where $\mathbf{J}(\mathbf{s})$ is the Jacobi matrix that maps from the task space to the synergy space, 
$\mathbf{r}$ is the residual of the task,
$\mathbf{G}_k$ is the Jacobi matrix of the $k$ constraint that transforms from the constraint space to the synergy space
and $\mathbf{h}_k$ is the residual of the $k$ constraint.
If the task is given in the synergy space, then $\mathbf{J}$ equals the identity matrix.
We solve this problem using one of the off-the-shelf quadratic programming solvers.

The Jacobi of the self-collision constraint $\mathbf{J}_{self}\!=\!\frac{\partial d}{\partial \mathbf{s}}$ is computed numerically,
where $d$ represents the penetration distance between two colliding links.
In order to speed up the collision checking computation, the meshes of the hand are modelled as capsules.
Because the task is considered in the cost function and the self-collision as a constraint,
in case of self-collisions,
the resulting grasps will approach as much as possible the target task without incurring in collisions.
Note that the IK solver is \textit{only} used to correct inferred grasp configurations (Sec.~\ref{sec:learning_synergies}) in case of self-collisions.
The motion interpolation and execution is achieved in joint space.

%% file: shape_space_reg.tex
\section{Shape Space Registration}
\label{sec:shape_space}
In this section we describe the shape space registration introduced before in \cite{Rodriguez2018a}.
We define a category as a set of objects with similar usage and extrinsic geometry.
Each object is represented as a point cloud.
A category contains a canonical model $\mathbf{C}$ which will be deformed towards the other objects of the category using CPD.
The shape space of the category is found by calculating the principal components of these deformations.

For two point sets, $\mathbf{Z}^{[t]} = (\mathbf{z}^{[t]}_1, ..., \mathbf{z}^{[t]}_M)^T$ and $\mathbf{Z}^{[r]} = (\mathbf{z}^{[r]}_1, ..., \mathbf{z}^{[r]}_N)^T$, CPD provides a deformation field that maps the points in $\mathbf{Z}^{[t]}$ into $\mathbf{Z}^{[r]}$.
For that, a Gaussian Mixture Model (GMM) is proposed such that the points in $\mathbf{Z}^{[t]}$ are considered centroids from which the points in $\mathbf{Z}^{[r]}$ are drawn. 
CPD maximizes the likelihood of the GMM while imposing constraints in the form of motion coherence on the centroids.
Points are only allowed to move coherently with the motion of their neighbors.
For the non-rigid case, CPD defines the transformation $\mathcal{T}$ from $\mathbf{Z}^{[t]}$ to $\mathbf{Z}^{[r]}$ as:
\begin{equation}
\mathcal{T}(\mathbf{Z}^{[t]},\mathbf{W}) = \mathbf{Z}^{[t]} + \mathbf{G}\mathbf{W}
\end{equation}
where $\mathbf{G}$ is a Gaussian kernel matrix such that $g_{ij}=G(\mathbf{z}^{[t]}_i, \mathbf{z}^{[t]}_j)=e^{-\frac{1}{2\beta^2}\|\mathbf{z}^{[t]}_i-\mathbf{z}^{[t]}_j\|}$, 
$\beta$ is a parameter that controls the influence between points and $\mathbf{W}$ is a matrix of coefficients. 
The matrix $\mathbf{W}$ is estimated in a Expectation Maximization fashion.
Please refer to \citep{Rodriguez2018a} or \cite{Rodriguez2018b} for further details.

We set the canonical model $\mathbf{C}$ as the deforming point set $\mathbf{Z}^{[t]}$ and each training sample $\mathbf{T}_i$ as the reference point set $\mathbf{Z}^{[r]}$, 
so the transformation $\mathcal{T}_i$ of each training sample $\mathbf{T}_i$ is described by:
\begin{equation}
\label{eq:defomation_field}
\mathcal{T}_i(\mathbf{C},\mathbf{W}_i) = \mathbf{C} + \mathbf{G}\mathbf{W}_i .
\end{equation}

From Eq. \ref{eq:defomation_field} we observe that the deformation is uniquely described by $\mathbf{W}_i$.
Note that $\mathbf{C}$ and $\mathbf{G}$ depend only on the canonical model and remain constant for all training samples.
In addition, the dimensionality of $\mathbf{W}_i\in\mathbb{R}^{M\times D}$ equals the dimensionality of the $\mathbf{C}\in\mathbb{R}^{M\times D}$,
which means that all matrices $\mathbf{W}_i$ can be organized such that elements in one matrix represent the same in another matrix.
This is a key feature for constructing the shape (latent) space.

For building the shape space, all matrices $\mathbf{W}_i$ are expressed as vectors and normalized to have unit-covariance and zero-mean.
Later, they are concatenated into an design matrix $\mathbf{Y}\in\mathbb{R}^{MD\times N}$.
We apply the Principle Component Analysis Expectation Maximization (PCA-EM) on $\mathbf{Y}$ to finally generate the shape space.
Thus, the shape of an instance can be described by a low dimensional $\mathbf{x}$ latent vector.
Note that PCA-EM also allows the transformation from the latent space to the deformation field manifold,
$\mathcal{W}$ will denote the function that performs such transformation.

For inferring the shape space of a new instance,
we search in the lower dimensional subspace to find a transformation which relates the canonical model to the observation at best.
We do this by optimizing a cost function using gradient descent.
Additionally, we incorporate a rigid transformation into the cost function, in order to account for small global misalignments.
An initial coarse alignment is required because of the numerous expected local minima.
Inspired by CPD, we optimize the following cost function:
\begin{equation}
\label{eq:Energy}
E(\mathbf{x},\bm{\theta}) = \sum^{M}_{m=1}{\sum^{N}_{n=1}{P\norm{\mathbf{O}_n-\Theta(\mathcal{T}_m(\mathbf{C}_m,\mathcal{W}(\mathbf{x})_m),\bm{\theta})}^2}} ,
\end{equation}
where $\Theta$ is a function that perform the rigid transformation given parameters $\bm{\theta}$ 
and $P$ represents the probability or importance weights between points expressed as:
\begin{equation}
	P = \frac{e^{\frac{1}{2\sigma^2}||\mathbf{O}_n-\Theta(\mathcal{T}(\mathbf{C}_m,\mathcal{W}(\mathbf{x})_m),\bm{\theta})||^2}}{\sum^{M}_{k=1}{{{e^{\frac{1}{2\sigma^2}||\mathbf{O}_n-\Theta(\mathcal{T}_k(\mathbf{C}_k,\mathcal{W}(\mathbf{x})_k),\bm{\theta})||^2}}}}} .
\end{equation}

\begin{figure}
	\centering
	\includegraphics[width=1.0\linewidth]{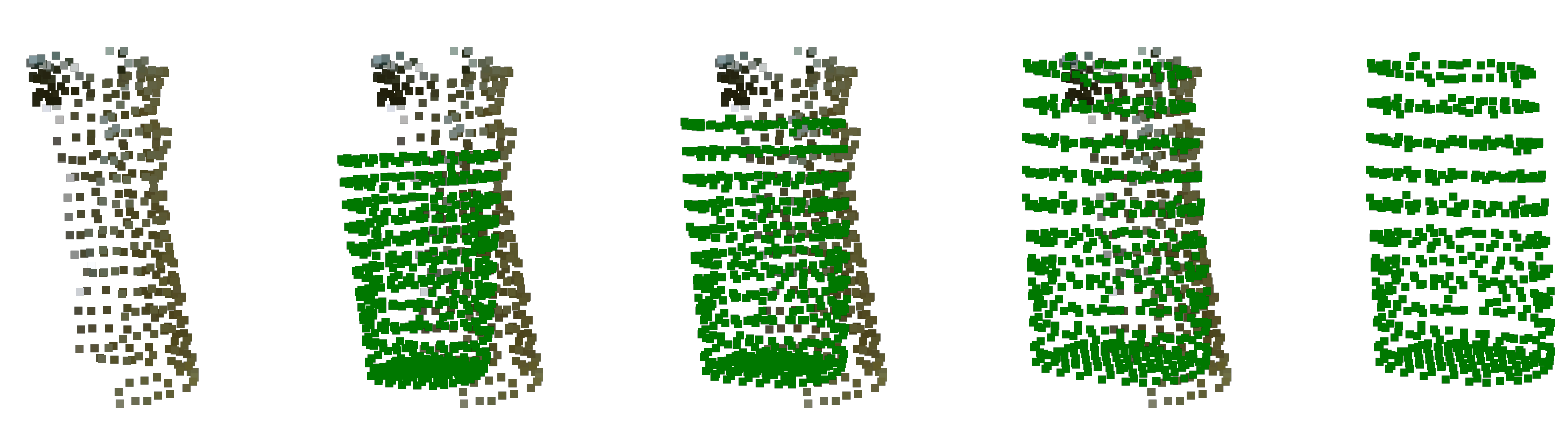} 
	\caption{Shape space registration of a glass used in the training phase for inferring the postural synergy.
		Observe how the object is reconstructed.}
	\label{fig:inference_glass}
\end{figure}

After convergence, the resulting vector $\mathbf{x}$ characterizes the shape of the observed instance.
The inference of a partially observed glass is shown in Fig.\ref{fig:inference_glass}.

\subsection{Automatic generation of object models}
\label{sec:aut_gen_models}
The number of available object models, i.e., the number of training samples, might limit the shape registration.
Even with the use of available on-line 3D object databases, the number of training instances might be small.
To overcome this limitation, we propose an automatic generation of object instances,
in which the canonical model is exposed to several constrained operations for generating new models.
These transformations include: global scale, $x$-, $y$-, $z$-, $xy$-, $xz$- and $yz$-scale, $xy$-, $xz$- and $yz$-projective transformations.
Each category defines a set of operations that fits with their geometry, 
e.g., the \textit{sphere} category will only apply a global scale to its canonical model to generate new instances.
The constraints are applied after all operations are performed in a consecutive manner.
The maximum dimensions, for instance, make part of these set of constraints.
The activation value of each operation is sampled from a multivariate Gaussian distribution parametrized considering typical values of the category's geometry.
If some of the constraints are not met, then the generated model is rejected.
After the generation process, some samples might still be removed by experts.
For objects with complex geometries that can be divided into parts, e.g. cylinder and handle for mugs, 
the operations can be applied individually to each part.
\begin{figure*}[]
	\centering
	\includegraphics[width=1.0\linewidth]{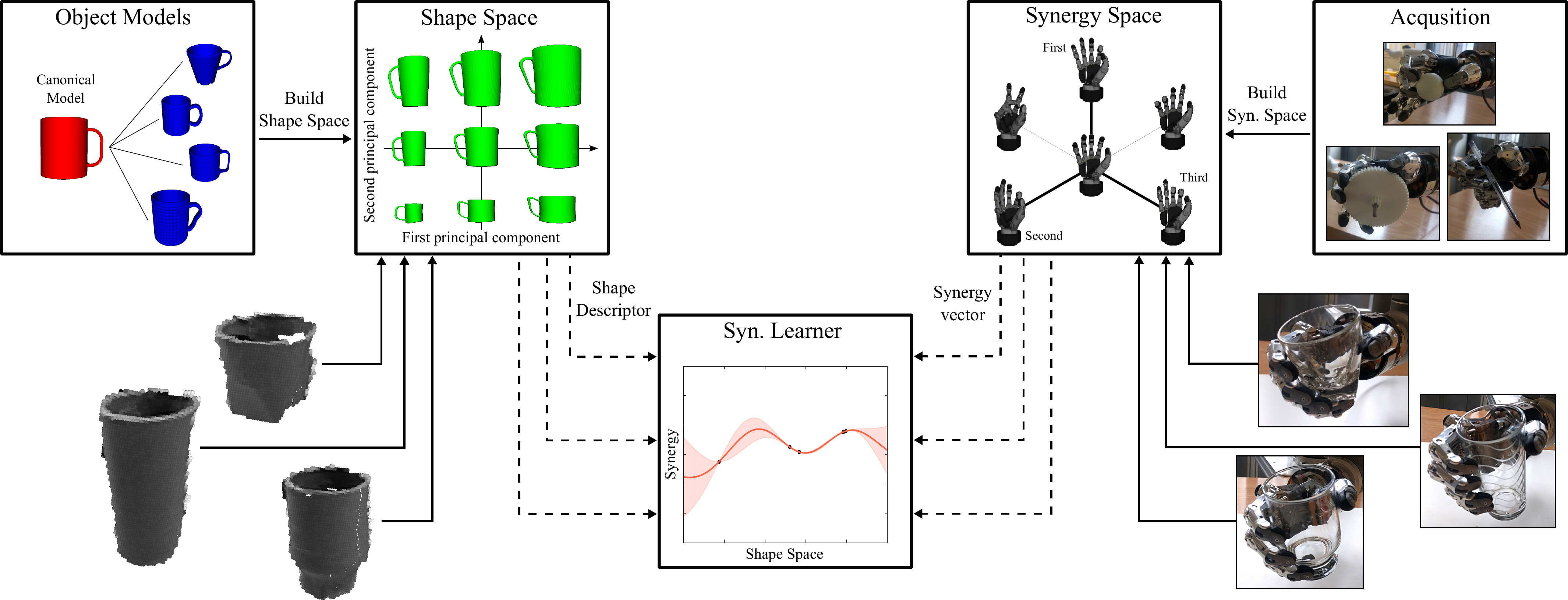} 
	\caption{Training the synergy learner. Initially, the shape space of the category and the synergy space of the robotic hand are calculated.
		Each of the training objects is then grasped to get their synergy vector. 
		Additionally, their respective shape descriptor is computed using the shape space registration.
		The synergy learner is finally trained by associating the shape descriptors with their respective synergy vectors.}
	\label{fig:trainig}
\end{figure*}
\begin{figure*}[]
	\centering
	\includegraphics[width=0.95\linewidth]{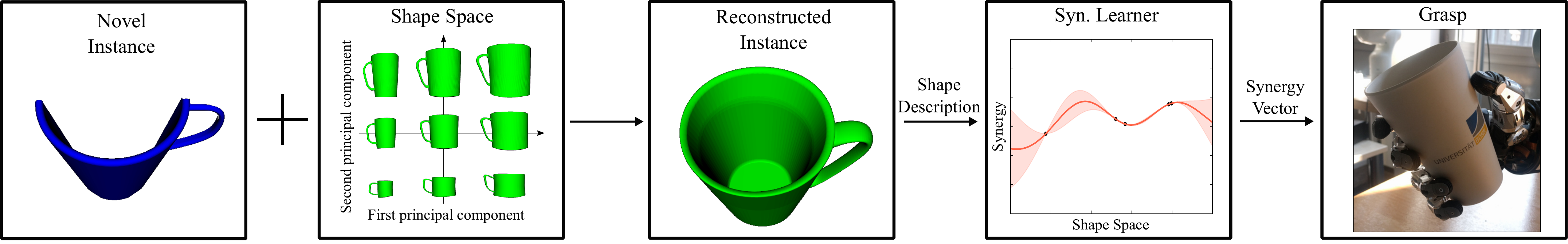} 
	\caption{Inference of synergy values. The observed instance is reconstructed by means of the shape space. The resulting shape descriptor is the input of the synergy learner which outputs the final grasping configurations as a synergy.}
	\label{fig:inference}
\end{figure*}

%% file: learn_synergies.tex
\section{Learning Postural Synergies}
\label{sec:learning_synergies}
We propose a supervised learning approach to learn grasps according to the shape of the objects.
The shape descriptor is the result of a non-rigid registration that incorporates category information as detailed in Sec.~\ref{sec:shape_space}.
On the other hand, the grasps are expressed by synergies.
Thus, the shape space of the category and the synergy space of the robotic hand have to be built before the training phase starts.

In the training phase, 
all the objects of the category are grasped by using the exoskeleton, 
and their respective synergy values are calculated.
Additionally, the shape descriptors of the same objects have to be computed using the shape registration.
For simple objects, such as spheres, the models can be created in simulation and converted into point clouds.
For more complex objects, real point clouds coming from 3D sensors are employed.
In this manner, synergy values of grasped objects belonging to a category are associated with their respective shape descriptor (Fig.~\ref{fig:trainig}).

Note that a single object can be grasped in several complete different manners,
i.e., different synergy vectors can be assigned to a single object shape descriptor.
We assume the synergy values associated to an object to be Gaussian distributed.
Thus, according to the number of synergies, several Gaussian Processes (GPs) are trained, one for each synergy.
In other words, each shape is mapped to a Gaussian distribution of grasps or synergy values.
All the GPs are parametrized with the Radial Basis Function kernel.

In the inference phase, given the shape descriptor, the synergy values are inferred from the mean predictions of the Gaussian Processes (Fig.~\ref{fig:inference}).
As a result of the shape registration, the observed model is reconstructed.

%% file: experiments.tex
\section{Experimental Results}
\label{sec:experiments}
\begin{figure*}
	\centering
	\includegraphics[width=1.0\linewidth]{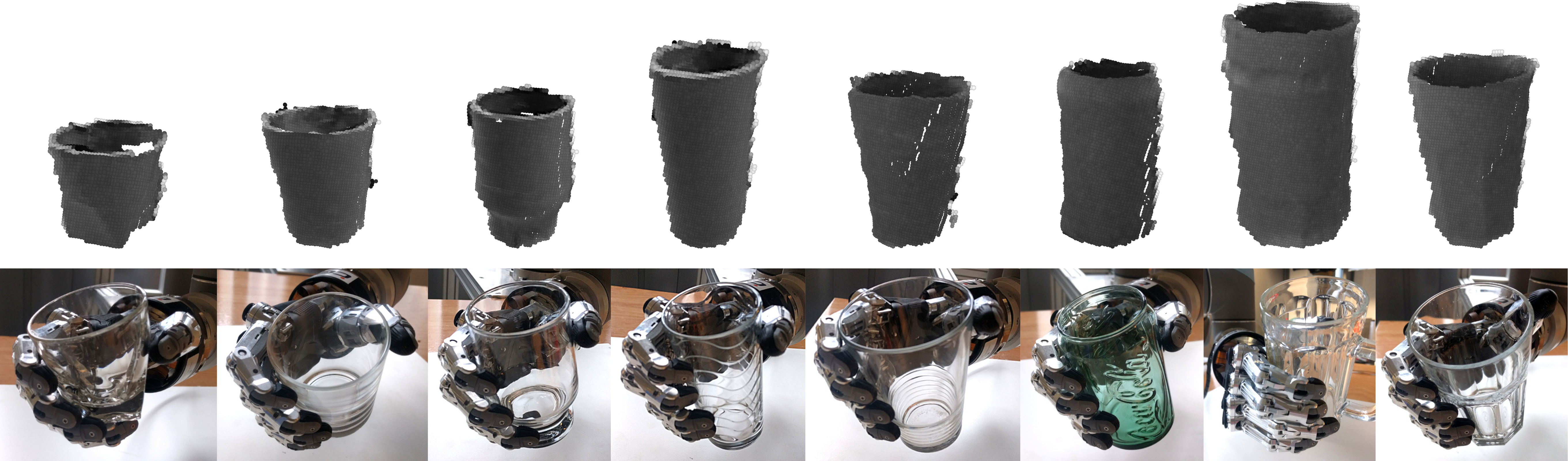} 
	\caption{Training instances for the synergy learner. Top: observed point clouds of the objects to be grasped, and bottom: the respective grasped performed through teleoperation.}
	\label{fig:training_grasps}
\end{figure*}
\begin{figure*}
	\centering
	\includegraphics[width=1.0\linewidth]{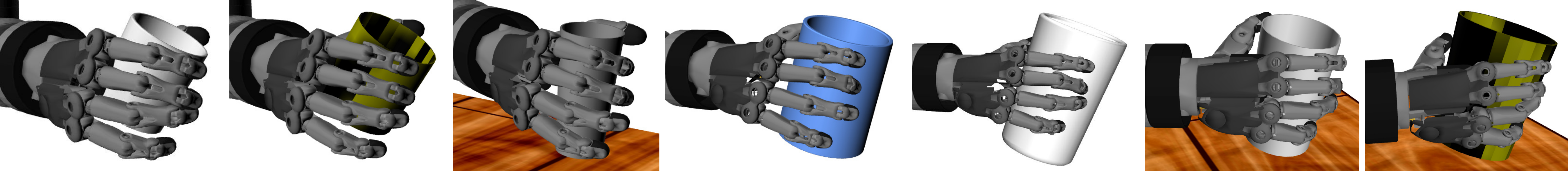} 
	\caption{Experiments performed in simulation. All the glasses were presented for the first and were successfully grasped. Note that, even though in some images a part of the table is displayed, there is no contact between it and the object. }
	\label{fig:gazebo_exp}
\end{figure*}
We tested our approach on two categories: \textit{Spheres} and \textit{Glasses}.
Models were obtained using 3D CAD databases.
Additionally, more object samples were added to the training set by using the auto-generation method as described in \ref{sec:aut_gen_models}.
Hence, the shape space of the categories was built only using data from virtual models.
The canonical model was selected by experts.
The object models were represented as point clouds,
which were obtained by ray-casting the meshes from several viewpoints on a tessellated sphere.
The resulting point cloud is down-sampled with a voxel grid filter.

The training sets for the shape space were composed of 9 spheres and 16 glasses.
Interestingly, the principal components found by our shape space registration coincides with our expectations.
For the spheres, the first (and only) component performs a global scale operation.
For the glasses, scale operations applied to the diameter, the height or both were found.
This reinforces the applicability of our shape space registration as a mean to describe object shapes.

For training the synergy learner, i.e., the Gaussian Processes $\mathcal{G}_i$ that are responsible for inferring the synergy values from the shape descriptor,
several objects were grasped using the same teleoperation scheme used for the generation of the synergy space.
The resulting joint configurations were transformed to the synergy space by Eq.~\ref{eq:to_synergy}.
The objects were perceived using the KinectV2 sensor~\cite{fankhauser2015}.
The raw 3D image of the sensor is filtered using a tabletop segmentation and a voxel grid filter to get a coarser point cloud.
The surfaces of the objects were slightly modified because of the difficulties of the sensor to perceive glass.
The grasped training objects together with their respective observed point clouds of the \textit{Glass} category are shown in Fig.~\ref{fig:training_grasps}.
Then, the shape descriptors were computed through the shape space registration.
Finally, each $\mathcal{G}_i$ was trained with the respective synergy values and shape descriptors.
\begin{figure}[b!]
	\centering
	\includegraphics[width=1.0\linewidth]{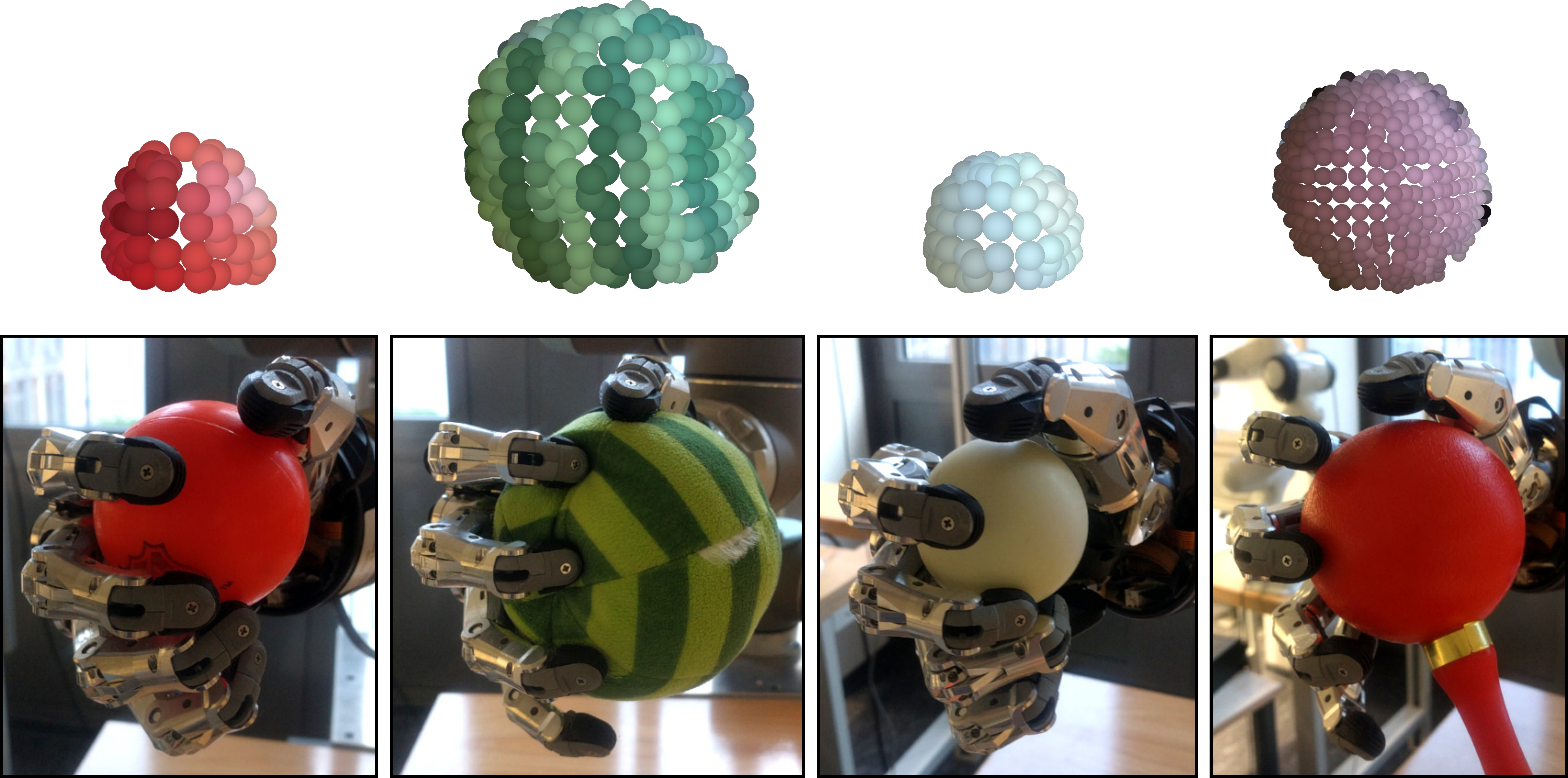} 
	\caption{Real robot experiments for the \textit{Sphere} category. 
		From left to right all three objects were successfully grasped.
		The last grasp failed to close enough the fingers.}
	\label{fig:real_balls}
\end{figure}

Initial evaluation of the inferred synergies was performed in physics-based simulations (Gazebo).
We attached the Schunk hand to a UR10 arm so that the objects can be reached.
The objects were placed in a known pose.
The manipulator approaches the object without making contact and slowly the hand moves toward the object such that a contact is guaranteed.
Note, nevertheless, that the approaching or pregrasp arm end-effector pose can be inferred as in previous works \cite{Rodriguez2018a, Rodriguez2018b}.
In the evaluation however these approaches were not employed in order to isolate and consequently to evaluate only the goodness of the inferred grasping configurations.
In the same manner, in order to know the object category and to estimate the pose, similar pipelines as in our related works, \cite{Klamt2018} and \cite{Pavlichenko2018}, can be integrated.
Thus, the grasp planning can be completely autonomous.
After the robotic hand reaches the inferred configuration, the arm tries to lift the object.
If the object does not fall for more than $10$ seconds after the lift motion finishes, the trial is counted as successful.
The testing set was composed of seven glasses, and all of them were successfully grasped.
Fig.~\ref{fig:gazebo_exp} shows the grasped objects.

Real robot experiments were also performed to evaluate our approach.
The robotic hand was controlled using a position-current cascade controller, 
For the $Sphere$ category, four objects of increasing radius were evaluated (Fig.~\ref{fig:real_balls}).
The objects were presented for the first time to the system.
The input point cloud is also shown to demonstrate that our method works with partial views.
Three of them were successfully grasped.
The grasp that failed is shown at the rightmost. % of Fig.~\ref{fig:real_balls}.
Qualitatively, the fingers were in the correct configuration but not close enough to establish the grasp.
We presume this was due to the few (six) training samples of the synergy learner.
Note that with an additional strategy to close the fingers until contact (current threshold), the grasp will succeed; 
however, we wanted to evaluate the inferred grasps purely.
In some grasps, the fingertips do not touch the object because of the mimic joints,
the contact is nevertheless ensured by the distal or proximal links.
\begin{figure*}
	\centering
	\includegraphics[width=1.0\linewidth]{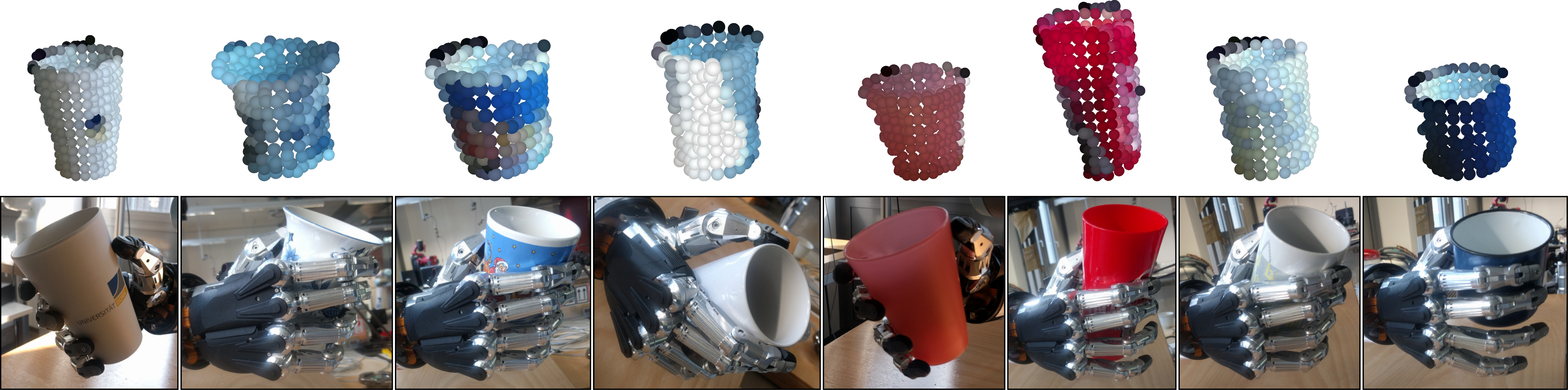} 
	\caption{Real robot experiments for the \textit{Glass} category. All the objects were presented for the first time to the system and they were successfully grasped.}
	\label{fig:real_glasses}
\end{figure*}

The $Glass$ category was also evaluated with the real robot.
Our approach was able to successfully grasp all eight novel objects (Fig.~\ref{fig:real_glasses}).
In average the inference time took $10\pm 0.8$ seconds, which confirms the applicability of this method in on-line grasping scenarios.

%% file: conclusion.tex
\section{Conclusion}
\label{sec:conclusion}
We have presented an approach for grasping novel objects belonging to a category based on their extrinsic geometry.
The effectiveness of this method was evaluated in both simulation and in real robot experiments.
The results showed that the representation of the geometry (coming from the shape space registration) and the representation of the grasp configuration (postural synergies) are good options for inferring grasps of novel objects.
One of the demonstrated advantages is the applicability in on-line scenarios.

For extracting postural synergies, the non-linear representation, GP-LVM, has shown lower reconstructions errors compared to its linear counterpart, especially for one and two dimensions~\cite{romero2013}.
In the future, we plan to evaluate the performance of applying GP-LVM instead of PCA.
Additionally, We also plan to evaluate the robustness of this method with more complex geometries.